\documentclass[letterpaper, 10 pt, conference]{ieeeconf}
\IEEEoverridecommandlockouts
\usepackage{graphicx} 
\usepackage{svg}
\usepackage{pgfplots}
\usepackage{float}
\usepackage{epsfig} 
\usepackage{mathptmx} 
\usepackage{times} 
\usepackage{amsmath} 
\usepackage{amssymb}  
\usepackage{bm}

\usepackage{array}
\usepackage{multirow}

\usepackage{tikz}

\usepackage{cite}

\pgfplotsset{compat=1.18}

\newlength{\cellwidth}
\divide\cellwidth by 4

\makeatletter
\let\NAT@parse\undefined
\makeatother
\usepackage{hyperref} 

\title{\LARGE \bf Theoretical Closed-loop Stability Bounds for Dynamical System Coupled with Diffusion Policies}

\author{Gabriel Lauzier, Alexandre Girard$^{1}$ and François Ferland$^{2}$
\thanks{$^{1}$Alexandre Girard is with the Department of Mechanical Engineering, Universite de Sherbrooke, Qc, Canada {\tt\small  alex.girard@usherbrooke.ca }}
\thanks{$^{2}$François Ferland is with the Department of Electronic and Computer Engineering, Universite de Sherbrooke, Qc, Canada {\tt\small  francois.ferland@usherbrooke.ca }}
}

\begin{document}

\maketitle
\thispagestyle{empty}
\pagestyle{empty}

\begin{abstract}

Diffusion policy has shown great performance in robotic manipulation tasks under stochastic perturbations, due to its ability to model multimodal action distributions. Nonetheless, its reliance on a computationally expensive reverse-time diffusion (denoising) process, for action inference, makes it challenging to use for real-time applications where quick decision-making is mandatory. This work studies the possibility of conducting the denoising process only partially before executing an action, allowing the plant to evolve according to its dynamics in parallel to the reverse-time diffusion dynamics ongoing on the computer. In a classical diffusion policy setting, the plant dynamics are usually slow and the two dynamical processes are uncoupled. Here, we investigate theoretical bounds on the stability of closed-loop systems using diffusion policies when the plant dynamics and the denoising dynamics are coupled. The contribution of this work gives a framework for faster imitation learning and a metric that yields if a controller will be stable based on the variance of the demonstrations.

\end{abstract}

\section{INTRODUCTION}

Robotic control and decision making increasingly rely on generative models to capture complex and multimodal action distributions. Vision-Language-Action (VLA) models \cite{kim_openvla_2024, zitkovich_rt2_2023} have demonstrated great performance in manipulation tasks and the interest in this field is growing with the uprising of large language models. The benefits of such models are the usage of pretrained models on large, Internet-scale datasets that allow the use of generic architecture \cite{kim_openvla_2024}. On the other hand, such datasets and training infrastructure are often hard to obtain, resulting in closed-source models hard to fine-tune for new platforms. Other works leveraging diffusion policies \cite{chi_diffusion_2024, chi_umi_2024} have shown efficient learning that requires fewer expert demonstrations, allowing someone to easily build a dataset for a given task. Despite its efficient learning, the hyperparameters of the model are often convoluted \cite{karras_elucidating_2022} and given the iterative nature of diffusion models, it requires several steps of inference, making them challenging to use for real-time applications.

In this work, we propose a \emph{partial diffusion policy} that executes only one step of the denoising diffusion process before taking action in the environment. Rather than expressing the system made of the plant dynamics and the denoising diffusion process as two decoupled systems, we analyze its stability as a coupled dynamical system. This allows us to show that the full denoising diffusion process from Diffusion Policy and the proposed partial diffusion policy are equivalent in stability depending on the variance of the expert demonstrations, the diffusion coefficient and the time-scale ratio between the plant dynamics and the controller.

The main contribution is a theoretical bound for asymptotic stability for a linear time-invariant plant under full and partial diffusion policies. This gives a metric for estimating the quality of a dataset given the variance of the expert demonstrations, the responses of the environment and the parametrization of the diffusion-based policy. It also shows a framework for trading off denoising depth against latency given proper variance in the expert demonstrations. This understanding of the diffusion-based policy brings such a framework closer to real-time robotics.

\section{RELATED WORKS}

\subsection{Score-Based Generative Modeling}
Song et al. \cite{song_score-based_2021} introduced a mathematical framework for denoising diffusion models that leverage stochastic differential equations (SDEs). Unlike likelihood-based methods, the suggested approach doesn't need to learn the partition function of the Boltzmann/Gibbs distribution by approximating the score function of the distribution with a neural network and a score matching objective \cite{hyvarinen_estimation_2005, vincent_connection_2011, song_generative_2019}. The process of learning and inference is defined respectively by two SDEs: the forward equation and the reverse-time equation:

\begin{equation} \label{eq:forward_time_sde}
    d\mathbf{x} = \mathbf{f}(\mathbf{x}, \tau)d\tau + \mathbf{g}(\mathbf{x}, \tau)d\mathbf{w}
\end{equation}

\begin{equation} \label{eq:reverse_time_sde}
    d\mathbf{x} = \left[\mathbf{f}(\mathbf{x}, \tau) - \mathbf{g}^2(\tau) s_\theta(\mathbf{x}, \tau)\right]d\tau + \mathbf{g}(\tau)d\hat{\mathbf{w}}
\end{equation}

where $\mathbf{w}$ is a standard Brownian motion, $\mathbf{f}$ is the drift coefficient, $\mathbf{g}$ is the diffusion coefficient and $\mathbf{s_\theta}$ is the approximation of the score function, the gradient of the log-probability. The reverse-time SDE in equation \ref{eq:reverse_time_sde} is then used to generate samples from the learned prior distribution. The discretization of variance-preserving SDE has been shown to yield Denoising Diffusion Probabilistic Models (DDPM) \cite{sohl-dickstein_deep_2015, ho_denoising_2020, song_score-based_2021}. 

\subsection{Diffusion Policy}
Diffusion Policy \cite{chi_diffusion_2024} applies a conditional DDPM model to learn a policy observed from expert demonstrations. This is done by learning the score function of the conditional distribution over actions. Then, at inference time, it iteratively optimizes the score function through a series of stochastic Langevin dynamics steps to generate a sequence of actions within a given horizon. In this work, the model is conditioned on visual observations.

By choosing to predict a sequence of actions instead of a single action, Diffusion Policy reduces jitter in the policy due to different valid modes between steps. It also allows handling idle actions where the agent shouldn't move, which is often difficult for single-step policies.

Another main advantage against other imitation learning methods is its capacity to model multimodal action distributions because of the stochastic nature of DDPM. This results directly in better performance with position control than with velocity control.

For a simple task in a linear dynamic system in standard state-space form where the demonstrations are given by a linear feedback policy: $\mathbf{u}_t = -\mathbf{K}\mathbf{x}_t$, the DDPM sampling will converge to the expert policy given a sequence of actions of length one. With a bigger horizon, this policy will converge to $\mathbf{u}_{t+t'} = - \mathbf{K}(\mathbf{A} - \mathbf{B}\mathbf{K})^{t'}\mathbf{x}_t$, which means that the policy must implicitly learn the dynamics of the system $\mathbf{A} - \mathbf{B}\mathbf{K}$.

\section{METHOD}
First, we define a linear time-invariant dynamic system in a standard state-space form:

\begin{equation}
    \mathbf{\dot{x}} = \mathbf{A}\mathbf{x} + \mathbf{B}\mathbf{u}
\end{equation}

Like Diffusion Policy, we aim to obtain demonstrations from a linear feedback policy: $\mathbf{u}_t = -\mathbf{K}\mathbf{x}_t$. Then we want a surrogate policy that imitates the expert behavior. Instead of using DDPM like Diffusion Policy, we use the score-based generative modeling framework \cite{song_score-based_2021} to model the denoising process. Equation \ref{eq:reverse_time_sde} is used for this purpose. Within this equation, the score function in the drift coefficient could be seen as a vector field that guides toward the learned prior probability density function. This last idea is what motivates this work. Instead of doing the full denoising process to sample an action, we directly take an action towards the direction of the most probable action. In a closed-loop system, this yields:

\begin{equation} \label{eq:closed-loop_system}
    \begin{bmatrix}
        d\mathbf{x} = \left[ \mathbf{A} \mathbf{x} + \mathbf{B} \mathbf{u} \right]dt \\
        d\mathbf{u} = \left[ \mathbf{f}(\mathbf{u}, \tau) + \mathbf{g}^2(\tau)\mathbf{s}_\theta(\mathbf{u}, \tau) \right]d\tau + \mathbf{g}(\tau)d\mathbf{\hat{w}_\tau}
    \end{bmatrix}
\end{equation}

Where $\tau$ if the diffusion time and the score function $\mathbf{s}_\theta(\mathbf{u}, \tau)$ is defined as the gradient of the log-probability of the expert policy with respect to the action $\mathbf{u}_t$:

\begin{equation}
    \mathbf{s}(\mathbf{u}) = \nabla_{\mathbf{u}} \log p(\mathbf{u})
\end{equation}

If the expert demonstration is deterministic, as we defined earlier, the score function will be ill-defined and could be interpreted as a Dirac delta function $\delta(\mathbf{u} - \mathbf{u}_0\exp(\mathbf{A} - \mathbf{B}\mathbf{K})t)$ which is a distribution that is zero everywhere except at the point $\mathbf{u} = \mathbf{u}_0\exp((\mathbf{A} - \mathbf{B}\mathbf{K})t)$. This means that anywhere other than the expert policy trajectory, there is no direction to go towards. This is a problem if we want to use the score function to guide the system towards the most probable action. To this matter, we model the expert policy as a stochastic policy using a Gaussian distribution centered on the deterministic policy:

\begin{equation}
    p(u) = \frac{1}{\sqrt{2\pi\sigma^2}}\exp\left(-\frac{(u + K x)^2}{2\sigma^2}\right)
\end{equation}

With this probability density function, we can compute the score function as:

\begin{equation}
    \mathbf{s}(\mathbf{u}) = -\frac{1}{\sigma^2}\left(\mathbf{u} + \mathbf{K} \mathbf{x}\right)
\end{equation}

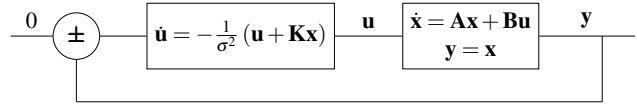
\begin{figure}
    \centering
    \resizebox{\columnwidth}{!}{%
        \tikzset{
  medium box/.style={
    fill=white,
    draw=black,
    rectangle,
    minimum width=0.75cm,
    minimum height=1cm,
    align=center
  },
  medium circle/.style={
    fill=white,
    draw=black,
    circle,
    minimum width=0.5cm,
    minimum height=0.5cm
  }
}

\begin{tikzpicture}
    \node [style=medium circle] (pm) at (-4.5, 0.0) {$\boldsymbol{\pm}$};

    \node [style=medium box] (controller) at (-2.0, 0.0) {$\dot{\mathbf{u}} = -\frac{1}{\sigma^2}\left(\mathbf{u} + \mathbf{K} \mathbf{x}\right)$};

    \node [style=medium box] (plant) at (1.5, 0.0) {$\dot{\mathbf{x}} = \mathbf{A}\mathbf{x} + \mathbf{B}\mathbf{u}$ \\ $\mathbf{y} = \mathbf{x}$};

    \coordinate (in) at (-5.5, 0.0);
    \coordinate (y) at (3.5, 0.0);
    \coordinate (y_end) at (4.0, 0.0);
    \coordinate (y_fb1) at (3.5, -1.0);
    \coordinate (y_fb2) at (-4.5, -1.0);

    \draw (in.center) to node [above] {$0$} (pm.west);
    \draw (pm.east) to (controller.west);
    \draw (controller.east) to node [auto] {$\mathbf{u}$} (plant.west);
    \draw (plant.east) to node [above, near end] {$\mathbf{y}$} (y.center);
    \draw (y.center) to (y_end.center);
    \draw (y.center) to (y_fb1.center);
    \draw (y_fb1.center) to (y_fb2.center);
    \draw (y_fb2.center) to (pm.south);
\end{tikzpicture}
    }
    \caption{Closed-loop dynamic system of a linear plant controlled by the denoising dynamics of a score-based generative model imitating a linear feedback policy.}
    \label{fig:lti_control}
\end{figure}

Notice that in figure \ref{fig:lti_control} instead of modeling the controller like in equation \ref{eq:closed-loop_system} we only use the score function. The effect of the drift function $\mathbf{f}(\mathbf{u}, t)$ is a translation of the solution of the equation. While the diffusion coefficient of the deterministic term $\mathbf{g}^2(t)$ could be seen as a gain on the step towards the most probable step. In the next subsection neither of them along with the stochastic term of the reverse-time SDE will be used to simplify the demonstration of stability. Also, note that since the score function $\mathbf{s}_\theta(\mathbf{u}, \tau)$ is dependent on the diffusion time $\tau$ but the expression of the stochastic policy isn't the joint probability of action and time, again, for simplicity. We will elaborate on the effect of the time diffusion and the diffusion coefficient on the stability in the subsection \ref{sec:effect_of_time}.

\subsection{Stability Analysis}
\subsubsection{1-Dimensional Case}

The system is stable if the eigenvalues are in the left half-plane. To this matter, we define an augmented system as:

\begin{equation*}
    \mathbf{x}^+ = \begin{bmatrix} x \\ u \end{bmatrix} \quad\quad\quad \lambda = \frac{1}{\sigma^2}
\end{equation*}

\begin{equation}
    \dot{x}^+ = \begin{bmatrix} \dot{x} \\ \dot{u} \end{bmatrix} = \begin{bmatrix} A x + B u \\ -\lambda K x - \lambda u \end{bmatrix} = \begin{bmatrix} A & B \\ -\lambda K & -\lambda \end{bmatrix} \begin{bmatrix} x \\ u \end{bmatrix}
\end{equation}

which gives the augmented state matrix $A^+$ as:

\begin{equation*}
    A^+ = \begin{bmatrix} A & B \\ -\lambda K & -\lambda \end{bmatrix}
\end{equation*}

By testing the inequality of the eigenvalues, we can see that the system is stable if the eigenvalues are in the left half-plane. This can be done by testing the following inequalities:

\begin{equation*}
    0 > A - \lambda \pm \sqrt{(A - \lambda)^2 - 4 \lambda (BK - A)} \\
\end{equation*}

which gives, by substituting for $\lambda$, the following inequalities:

\begin{equation} \label{eq:lti-proportional-stability}
    0 > A - BK
\end{equation}

\begin{equation} \label{eq:lti-svfg-stability}
    \sigma < \frac{1}{\sqrt{A}}, \quad\quad A > 0
\end{equation}

Where equation \ref{eq:lti-proportional-stability} is simply the stability of an LTI system under a linear feedback control, and where equation \ref{eq:lti-svfg-stability} is the stability condition related to the variance of the expert demonstrations. This means that the score-based closed-loop control is unstable if the demonstrator is unstable, and when the demonstrator is stable, the learned policy is only stable if the variance of the demonstrations is smaller than the inverse response of the system. This satisfies the hypothesis that the system is as good as the expert policy, given the quality of the demonstrations.

\begin{table}[h]
\centering
\renewcommand{\arraystretch}{1.3}
\caption{Stability bounds of the closed-loop linear time-invariant dynamics coupled with a diffusion policy.}
\label{table:stability_bounds}
    \begin{tabular}{m{0.8\cellwidth}|m{1.4\cellwidth}|m{\cellwidth}|m{0.8\cellwidth}}
        \hline
        Natural \newline Dynamic & Closed-Loop \newline Dynamic & Diffusion \newline gain & Stability \\
        \hline\hline
        \multirow[c]{2}{\cellwidth}{$\mathbf{A} > 0$}
                     & $\mathbf{A}-\mathbf{B}\mathbf{K} \geq 0$ & for all $\mathbf{K}'$ & unstable \\ \cline{2-4}
                     & $\mathbf{A}-\mathbf{B}\mathbf{K} < 0$ & $\mathbf{K}' \geq \mathbf{A}$ & stable \\
        \hline
        $\mathbf{A} < 0$ & for all $\mathbf{A}-\mathbf{B}\mathbf{K}$ & $\mathbf{K}' \geq 0$ & stable \\
        \hline
    \end{tabular}
\end{table}

\subsubsection{N-Dimensional Case}
The same idea applies to the N-dimensional case, except that instead of computing the eigenvalues of the state matrix in the state representation, we can elevate the system to a second-order ordinary differential equation and show that the second and third coefficients are positive definite, such as a matrix $\mathbf{M}$ with respect to $\mathbf{x}^\top\mathbf{M}x > 0$ for all $\mathbf{x} \in \mathbb{R}^N$ where $\mathbf{x} \neq 0$:

\begin{equation}
    \mathbf{\ddot{x}} - \left[\mathbf{A} - \Sigma^{-1} \right] \mathbf{\dot{x}} - \Sigma^{-1} \left[ \mathbf{A} - \mathbf{B} \mathbf{K} \right] \mathbf{x} = 0
\end{equation}

\begin{equation*}
    - \left[\mathbf{A} - \Sigma^{-1} \right] \succ 0 \quad\text{and}\quad -\Sigma^{-1} \left[ \mathbf{A} - \mathbf{B} \mathbf{K} \right] \succ 0
\end{equation*}

Where $\Sigma$ is the covariance matrix of the demonstrations, which is assumed to be linearly independent. 

We know that a matrix $\mathbf{M}$ is positive definite if and only if all its eigenvalues are positive given that $\mathbf{M}$ is symmetric. Since our matrix might not be symmetric, we can use only its symmetric part $\mathbf{M}' = \frac{1}{2}(\mathbf{M} + \mathbf{M}^\top)$ and show that the eigenvalues of $\mathbf{M}'$ are positive, as said earlier. With some simplification shown in the appendix \ref{appendix:extended_n_dimension_stability_bounds}, we show that the eigenvalues of the symmetric state matrix $\mathbf{A}$ and the symmetric state matrix under linear feedback control $\mathbf{A}-\mathbf{B}\mathbf{K}$ must respect the following inequalities given isotropic distributions of the demonstrations:

\begin{equation} \label{eq:ndim-lti-svfg-stability}
    \sigma < \frac{1}{\sqrt{\lambda_{\text{max}}(\mathbf{S_1})}}
\end{equation}

\begin{equation} \label{eq:ndim-lti-svfg-stability2}
    \lambda_{\text{max}}(\mathbf{S_2}) < 0
\end{equation}

For $\lambda_{\text{max}}(\mathbf{S_1})$ being the maximum eigenvalue of the symmetric matrix $\frac{1}{2}(\mathbf{A} + \mathbf{A}^\top)$ and $\lambda_{\text{max}}(\mathbf{S_2})$ being the maximum eigenvalue of the symmetric matrix $\frac{1}{2}(\left (\mathbf{A} - \mathbf{BK} \right) + \left (\mathbf{A} - \mathbf{BK} \right)^\top)$. This result matches the 1-dimensional case shown earlier.

\subsection{The Effect of Different Time Scale} \label{sec:effect_of_time}
Earlier we made the assumption that the derivative of the control law $\mathbf{\dot{u}}$ was made only with the score function at the same time scale as the plant ($dt = d\tau$). In its original work, Diffusion Policy uses a different time-scale and performs the action dynamics in an inner loop within plant dynamics ($dt \neq d\tau$). In this section, we show that the variance, the diffusion coefficient and different time scale act as a proportional gain applied to the error between the current action and the expert demonstration. Using the chain rule, we can rewrite the action dynamics as follows:

\begin{equation*}
    \frac{d\mathbf{u}}{dt} = \frac{d\tau}{dt}\frac{d\mathbf{u}}{d\tau} = \alpha \frac{d\mathbf{u}}{d\tau}
\end{equation*}

\begin{equation}
    d\mathbf{u} = \left[ \mathbf{f}(\mathbf{u}, \alpha t) + \mathbf{g}^2(\alpha t)\mathbf{s}_\theta(\mathbf{u}, \alpha t) \right]\alpha dt + \sqrt{\alpha}\mathbf{g}(\alpha t)d\mathbf{\hat{w}_t}
\end{equation}

Where the standard Brownian motion $d\mathbf{\hat{w}_\tau} = \sqrt{d\tau} \mathcal{N}(0, 1) = \sqrt{\alpha dt}\mathcal{N}(0, 1) = \sqrt{\alpha}d\mathbf{\hat{w}_t}$. By distributing $\alpha$ we see that different time scales have the effect of a gain on the drift coefficient $\mathbf{f}(\mathbf{u}, \alpha t)$ and on the score function $\mathbf{s}_\theta(\mathbf{u}, \alpha t)$. As said earlier, a non-zero drift coefficient doesn't affect the stability of the system and only translate the solution of the system of equations. Together, the variance, the diffusion coefficient and the time scale ratio between the plant time and the diffusion time are analogous to a proportional gain on the error of the current action such as $-\mathbf{K}'(\mathbf{u} + \mathbf{K}\mathbf{x})$. If we ignore the stochastic term, such action dynamics yield this condition of stability:

\begin{equation} \label{eq:lti-svfg-general-stability}
    \sigma < \mathbf{g}\sqrt{\frac{\alpha}{A}}
\end{equation}

\begin{figure}
    \centering
    \input{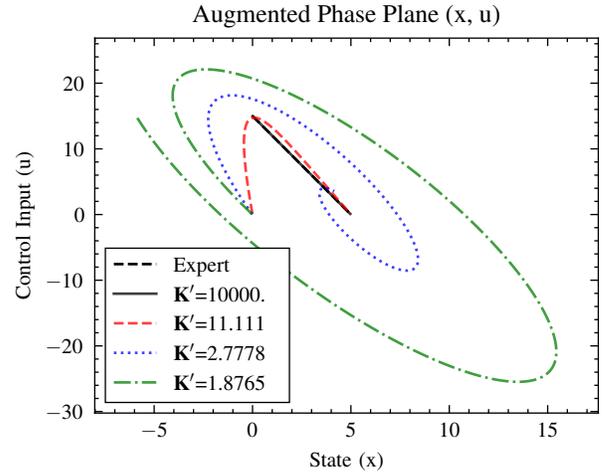}
    \caption{Trajectories of the expert policy and the score-based controller going from $x = 0$ to $x = 5$ under different gain $\mathbf{K}'$ in the augmented phase plane of a closed-loop system with a linear dynamics plant parametrized with $A = 2, B = 1, K = 3$. Given these parameters and the equation \ref{eq:lti-svfg-general-stability} we can see that the system starts being unstable when $\mathbf{K}' < A$.}
    \label{fig:augmented_phase_plane}
\end{figure}

\begin{figure}
    \centering
    \input{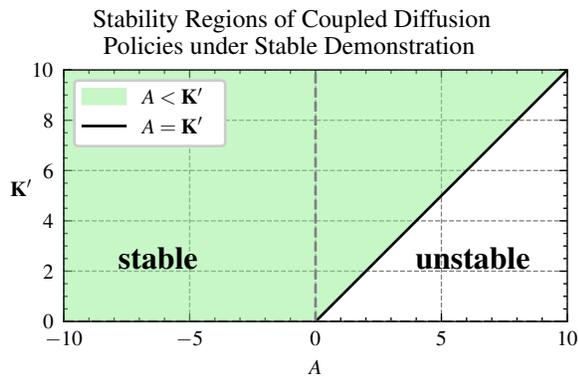}
    \caption{Region of stability of the coupled diffusion policies under stable demonstration $A - BK < 0$ where $K' = \mathbf{g}^2\alpha / \sigma^2$.}
    \label{fig:stability_region}
\end{figure}

With that in mind, figure \ref{fig:augmented_phase_plane} shows the augmented phase plane with $\mathbf{K}'$ being parametrized by the variance, the diffusion coefficient and the time-scale ratio such as $\mathbf{K}' = \mathbf{g}^2\alpha / \sigma^2$. The figure \ref{fig:stability_region} expresses the stability region of the diffusion policies given by equation \ref{eq:lti-svfg-general-stability}.

\section{DISCUSSION}

The methods shown allow the formulation of a theoretical bound in equation \ref{eq:lti-svfg-general-stability} that can be used to estimate the quality of an imitation learning dataset from the variance in the dataset given the response of the system and the parameterization of the diffusion policy framework. This gives the insight that given a fast-responsive system, the demonstrations should have lower variances and, on the other hand, given a slow responsive system, the demonstrations could have a higher variance and still guarantee the stability of the controller.

The methods also show that a per-step coupling of the Diffusion Policy against its classical multi-step coupling will display the same condition of stability. Diffusion Policy is implemented as an inner loop of the plant dynamics. Albeit being the inner loop, in real world implementation, it doesn't mean that one or multiple step of the Diffusion Policy is faster than the environment time scale. In practice, the denoising diffusion process is often slower than the dynamics of the environment. It's even more true given high-dimensionality input like an image as condition, which takes more time to process. This is the main limitation of these theoretical bounds for stability, which assume enough time to fit the computation of the action within the dynamics time scale of the plant. Even with this limitation, the per-step coupling method reduces the computation load compared to the multi-steps one, allowing faster inference of the actions. This makes it the closest method that could satisfy the theoretical bounds in real life scenarios.

\section{CONCLUSION}

The recent success of diffusion policies with manipulation tasks makes wonder if it's possible to apply such framework to real-time task like autonomous driving. At first sight, the iterative nature of the diffusion policies inference seems not appropriate for such applications. Here, we explore the stability condition of a linear-time invariant system coupled with a diffusion policy controller. The coupling of both systems allows showing the equality in stability of a partial diffusion policy against a fully iterative diffusion policy, which is bound by the equation \ref{eq:lti-svfg-general-stability}. This last equation can be used as a quantitative measure of the quality of a dataset given known parameters such as the variance of the expert demonstration, the diffusion coefficient and the time-scale ratio. In last, by showing the equality in stability of a partial diffusion policy, it allows the design of faster inference process with fewer steps.

Looking forward, several extensions to this work could be made. First, this work only takes into account the deterministic behavior of the diffusion policy system. Adding stochasticity to the system will affect the stability bounds. Future work should study the effect of such perturbations on the coupled system. Second, this work could be extended to nonlinear environment and analysis using Lyapunov stability theory, which could result in stability bound conditions generalizable to a broader range of applications. Finally, validation on high-dimensional space such as vision-conditioned tasks would allow a comparison with the original Diffusion Policy and quantify the gain and trade off of using a partial diffusion policy.

\bibliographystyle{IEEEtran}
\bibliography{references}

\appendices

\section{Extended Derivation of the N-Dimensional Stability Bounds} \label{appendix:extended_n_dimension_stability_bounds}
Below is the full derivation of \ref{eq:ndim-lti-svfg-stability} and \ref{eq:ndim-lti-svfg-stability2}, the stability bounds of a closed-loop linear time-invariant dynamics coupled with a diffusion policy. We first elevate the system in state-space form to a second-order ordinary differential equation:

\begin{equation}
    \mathbf{\ddot{x}} = \mathbf{A} \mathbf{\dot{x}} + \mathbf{B} \mathbf{\dot{u}}, \quad \mathbf{\dot{u}} = -\Sigma^{-1}(\mathbf{u} + \mathbf{K} \mathbf{x}).
\end{equation}

Where $\mathbf{\dot{u}}$ is the score function and $\Sigma$ is the covariance of the demonstrations, which we assume to be linearly independent. With $\mathbf{u} = \mathbf{B}^{-1} \left(\mathbf{\dot{x}} - \mathbf{A} \mathbf{x} \right)$ obtained from the state-space form, which gives:

\begin{align*}
    \mathbf{\ddot{x}} = \mathbf{A} \mathbf{\dot{x}} + \mathbf{B} \mathbf{\dot{u}} &= \mathbf{A} \mathbf{\dot{x}} + \mathbf{B} \left( -\Sigma^{-1} \mathbf{u} - \Sigma^{-1} \mathbf{K} \mathbf{x} \right) \\
    &= \mathbf{A} \mathbf{\dot{x}} - \Sigma^{-1} \mathbf{B} \mathbf{K} \mathbf{x} - \Sigma^{-1} \mathbf{B} \mathbf{u} \\
    &= \mathbf{A} \mathbf{\dot{x}} - \Sigma^{-1} \mathbf{B} \mathbf{K} \mathbf{x} - \Sigma^{-1} \mathbf{B} \mathbf{B}^{-1} \left(\mathbf{\dot{x}} - \mathbf{A} \mathbf{x} \right) \\
    &= \mathbf{A} \mathbf{\dot{x}} - \Sigma^{-1} \mathbf{B} \mathbf{K} \mathbf{x} - \Sigma^{-1}\mathbf{\dot{x}} + \Sigma^{-1} \mathbf{A} \mathbf{x} \\
\end{align*}

\begin{equation}
    \mathbf{\ddot{x}} - \left[\mathbf{A} - \Sigma^{-1} \right] \mathbf{\dot{x}} - \Sigma^{-1} \left[ \mathbf{A} - \mathbf{B} \mathbf{K} \right] \mathbf{x} = 0.
\end{equation}

To be stable, the second and third coefficients of this second-order system must be positive definite, such as a matrix $\mathbf{M}$ with respect to $x^\top\mathbf{M}x > 0$ for all $x \in \mathbb{R}^N$, where $x \neq 0$. This leads to the following condition:

\begin{equation}
    - \left[\mathbf{A} - \Sigma^{-1} \right] \succ 0 \quad\text{and}\quad -\Sigma^{-1} \left[ \mathbf{A} - \mathbf{B} \mathbf{K} \right] \succ 0.
\end{equation}

We know that $\mathbf{M}$ is positive definite if and only if all its eigenvalues are positive given that $\mathbf{M}$ is symmetric. Since our matrix might not be symmetric, we can use only its symmetric part $\mathbf{M}' = \frac{1}{2}(\mathbf{M} + \mathbf{M}^\top)$ and show that the eigenvalues of $\mathbf{M}'$ are positive.

\subsection{Positive definiteness of \texorpdfstring{$-\left[\mathbf{A} - \Sigma^{-1} \right]$}{the second term's coefficient}}

The symmetric part of the matrix of the coefficient $-\left[\mathbf{A} - \Sigma^{-1} \right]$ is given by

\begin{equation}
    -\frac{1}{2}(\mathbf{A} + \mathbf{A}^\top) + \Sigma^{-1} = \Sigma^{-1} - \mathbf{S_1}
\end{equation}

Where $\mathbf{S_1}$ is the symmetric part of the matrix $\mathbf{A}$ used for simplicity. Hence, we are interested in the condition $\Sigma^{-1} - \mathbf{S_1} \succ 0$. Which could be satisfied if:

\begin{equation}
    \lambda_{\text{min}}(\Sigma^{-1}) > \lambda_{\text{max}}(\mathbf{S_1}).
\end{equation}

Where $\lambda(\cdot)$ is the eigenvalues of the given matrix. In the special case of an isotropic distribution where $\Sigma = \sigma^2 \mathbb{I}$ we retrieve an inequality similar to the one-dimensional case in equation \ref{eq:lti-svfg-stability}.

\begin{equation*}
    \sigma < \frac{1}{\sqrt{\lambda_{\text{max}}(\mathbf{S_1})}}
\end{equation*}

\subsection{Positive definiteness of \texorpdfstring{$-\Sigma^{-1} \left[ \mathbf{A} - \mathbf{B} \mathbf{K} \right]$}{the second term's coefficient}}

The symmetric par of the matrix of the coefficient $-\Sigma^{-1} \left[ \mathbf{A} - \mathbf{B} \mathbf{K} \right]$ is given by:

\begin{equation}
    -\frac{1}{2}\left(\Sigma^{-1}\left(\mathbf{A} - \mathbf{B}\mathbf{K} \right) + \left(\mathbf{A} - \mathbf{B}\mathbf{K} \right)^\top\Sigma^{-1}\right)
\end{equation}

Therefore, the exact condition for stability is

\begin{equation}
    -\frac{1}{2}\left(\Sigma^{-1}\left(\mathbf{A} - \mathbf{B}\mathbf{K} \right) + \left(\mathbf{A} - \mathbf{B}\mathbf{K} \right)^\top\Sigma^{-1}\right) \succ 0.
\end{equation}

Which is equivalent to the Lyapunov inequality

\begin{equation}
    \Sigma^{-1}\left(\mathbf{A} - \mathbf{B}\mathbf{K} \right) + \left(\mathbf{A} - \mathbf{B}\mathbf{K} \right)^\top\Sigma^{-1} \prec 0.
\end{equation}

If all real parts of the eigenvalues of $\mathbf{A} - \mathbf{B}\mathbf{K}$ are strictly negative and since the covariance matrix $\Sigma$ is positive definite, then there exists a covariance matrix that meets this inequality.

In the special case of an isotropic distribution where $\Sigma = \sigma^2 \mathbb{I}$ we get

\begin{equation*}
    \left(\mathbf{A} - \mathbf{B}\mathbf{K} \right) + \left(\mathbf{A} - \mathbf{B}\mathbf{K} \right)^\top \prec 0.
\end{equation*}

Which is equivalent to the positive definiteness of the symmetric part of the matrix $\mathbf{A} - \mathbf{B}\mathbf{K}$ defined as $\mathbf{S_2}$ times the inverse of the covariance matrix:

\begin{equation}
    \frac{-\Sigma^{-1}}{2}\left(\left(\mathbf{A} - \mathbf{B}\mathbf{K}\right) + \left(\mathbf{A} - \mathbf{B}\mathbf{K}\right)^\top\right) = -\Sigma^{-1} \mathbf{S_2}
\end{equation}

Hence, we are interested in the condition $\Sigma^{-1} \mathbf{S_2} \succ 0$ which could be satisfied, in the isotropic case, if:

\begin{equation}
    \lambda_{\text{max}}(\mathbf{S_2}) < 0
\end{equation}

\end{document}